\documentclass[11pt]{article}

\PassOptionsToPackage{table}{xcolor}
\usepackage[preprint]{acl}

\usepackage{times}
\usepackage{latexsym}
\usepackage[T1]{fontenc}
\usepackage[utf8]{inputenc}
\usepackage{microtype}
\usepackage{inconsolata}

\usepackage{graphicx}
\usepackage{xcolor}
\usepackage{tabularx}
\usepackage{booktabs}
\usepackage{multirow}
\usepackage{array}
\usepackage{amsmath}
\usepackage{amsfonts}
\usepackage{nicefrac}
\usepackage{subcaption}
\usepackage[ruled,vlined,linesnumbered]{algorithm2e}

\SetKwInput{KwIn}{Input}
\SetKwInput{KwOut}{Output}
\DontPrintSemicolon
\SetAlgoNlRelativeSize{-1}
\SetAlCapSkip{0.5em}
\title{GeoReform: Reflective Formalization Evolution for Multimodal Geometry Problem Solving}

\author{
  Jialu Wang\textsuperscript{1},
  Ruichen Zhang\textsuperscript{2},
  Xiaoou Liu\textsuperscript{3},
  Hua Wei\textsuperscript{3},
  Tianlong Chen\textsuperscript{2,\dag} \\
  \textsuperscript{1}Tongji University \quad
  \textsuperscript{2}University of North Carolina at Chapel Hill \quad
  \textsuperscript{3}Arizona State University \\
  \textsuperscript{\dag}Corresponding author: \texttt{tianlong@cs.unc.edu}
}

\begin{document}

\maketitle 

% 1. example
% 在。。。case下会做得不好 verbally概括

% 2. MLLM unacpable to 提取and model 精细的几何结构，only from rawand complicated input。 which is 常见的paragram in ，，，。
% 中文意思写英文 这句话主谓宾胃 首先中文讲清楚

% 一会说重要 一会不好
% to address前 ， 怎么更好的表达这个数学问题 引入 in b 。zero-shot 表现 expecilly 显得formalization的范式 convert raW geometry problem into structed 怎么个representation。 

% 这个formalization是highly non-trival 怎么去orgenize这个structure会直接影响performance 不恰当的会afferct

% to fully 利用the power ，design一个 reflect，， navily adjust forma，which is xx compared to initial。有多少点的提升

% 和jbowledge disstill之间的关联 
% 直接和gpt5比 

% 衔接 做第二个的目的是第一个很sensitive

% 第一阶段sft一下

% GEPApaper的实验setting参考
% // 把我们自己的方法在不同开源/闭源model进行利用
% 对比不用我们方法的性能（其他方法的数

\begin{abstract}
Multimodal large language models (MLLMs) often struggle to identify and use geometric relations in diagrams.
Recent methods address this challenge by converting geometric entities, relations, and constraints into explicit textual representations for the model to reason over.
However, effective formalization is highly non-trivial: on Geometry3K, structure injection fixes 28 errors but introduces 13 new ones among 200 examples. Redundant relations can distract the model, while ambiguous references to diagram elements can lead it to apply constraints incorrectly. This suggests that the key challenge is not merely extracting more geometric facts, but organizing them into representations that support downstream reasoning. 
To fully exploit the power of formalization, we further propose GeoReform, a reflective formalization evolution framework that treats formalization as an optimizable policy rather than a fixed parser output. 
GeoReform executes the full reasoning pipeline, collects failed rollouts, diagnoses defects in the current representation, and mutates the policy to better select, ground, group, and present geometric entities, relations, constraints, and targets. 
On Geometry3K, GeoReform improves Qwen3VL-2B accuracy from 42.0\% to 56.0\%. 
Extensive experiments and analyses across geometry reasoning benchmarks demonstrate that effective formalization is crucial for improving multimodal geometry reasoning.
\end{abstract}

% Geometry problem solving (GPS) requires both accurate diagram understanding and rigorous mathematical reasoning. 
% Despite their strong performance on general tasks, multimodal large language models (MLLMs) often struggle to extract and model fine-grained geometric entities and relations from raw and complex multimodal inputs, which remains a common failure mode in GPS. 
% We show that explicitly formalizing geometry problems into structured representations can substantially improve reasoning. 
% However, effective formalization is highly non-trivial: the way geometric information is organized can significantly affect reasoning performance, while poorly designed formalizations can even degrade reasoning accuracy. 
% To fully exploit the power of formalization, we further introduce a reflective formalization evolution framework that refines the organization of geometric entities, relations, and constraints using feedback from downstream reasoning failures, improving accuracy from 42.0\% to 53.0\% on Geometry3K. 
% Extensive experiments and analyses on benchmarks demonstrate that effective formalization is crucial for improving multimodal geometry reasoning.
% \end{abstract}

% 删掉引入parser in MLLM

% 发现effective formalization是selective 多余的geomrtrics facts可以干扰reasoning （证据）
% multi-seed
% 59 +- 0.4

\section{Introduction}

Geometry problem solving (GPS) requires both accurate understanding of diagrams and rigorous mathematical reasoning~\citep{seo2015solving, chen2021geoqa, lu2021intergps, chen2022unigeo, zhang2024geoeval, lu2024mathvista, zhang2024mathverse, qiao2025wemath}. 
Recent multimodal large language models (MLLMs) have shown strong performance on a wide range of vision and language tasks, including visual question answering, instruction following, and multimodal reasoning~\citep{alayrac2022flamingo, li2023blip2, dai2023instructblip, liu2023visual, bai2023qwenvl, zhu2024minigpt4}. 
However, their success on general tasks does not directly translate to reliable geometry reasoning. MLLMs often struggle to extract and model detailed geometric entities and relations, such as points, segments, angles, perpendicular lines, and equality constraints~\citep{gao2025gllava, xia2025geox, ning2025gns, chen2025geotinyllava}.

\begin{figure*}[t]
    \centering
    \includegraphics[width=\textwidth]{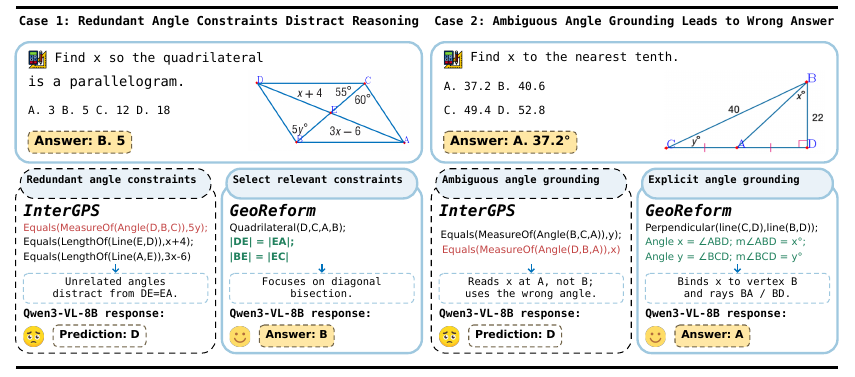}
   \caption{
Failure modes of naive geometry formalization. Left: redundant angle relations distract the model from the key segment relation. Right: unclear angle grounding causes the model to apply the wrong angle constraint.
}
    \label{fig:mathfailure}
\end{figure*}

To address these difficulties, recent GPS methods explicitly represent geometric entities, relations, and constraints and provide them to the reasoning model~\citep{gao2025gllava, xia2025geox, ning2025gns, chen2025geotinyllava}. However, we observe that effective formalization is highly non-trivial. 
Although a clear structure can guide MLLMs toward relevant entities, relations, and constraints~\citep{lu2021intergps, chen2021geoqa, chen2022unigeo, zhang2023pgpsnet, xia2025geox}, poorly designed formalizations can also interfere with the reasoning process. On 200 Geometry3K problems~\citep{lu2021intergps}, adding explicit geometric entities, relations, and constraints to the input of Qwen3VL-8B corrects 28 previously incorrect answers but turns 13 previously correct answers into errors.

We further observe that such interference occurs in two common ways. 
First, redundant or weakly relevant relations can distract the reasoning from the simplest solution path. 
As shown in Figure~\ref{fig:mathfailure} Case 1, the problem can be solved directly using the fact that the diagonals of a parallelogram bisect each other, but the formalized input introduces additional angle relations that steer the reasoning away from the key segment equality. 
Second, ambiguous formalizations can be amplified by downstream reasoning. As shown in Figure~\ref{fig:mathfailure} Case 2, angle relations with unclear diagram grounding can cause the model to apply constraints to the wrong angles and arrive at an incorrect answer. These failures reveal a gap between describing a diagram and representing it for problem solving.
Even valid geometric relations can distract reasoning when they are poorly selected or presented.
Effective formalization therefore requires deciding which information to include and how to express it, in addition to extracting geometric facts accurately.
This motivates using downstream reasoning feedback to improve the policy that governs these decisions.

To better exploit formalization, we propose GeoReform, a reflective formalization evolution framework that optimizes the formalization policy rather than treating parser outputs as fixed. 
GeoReform searches over candidate policies that determine how geometric entities, relations, constraints, and targets are selected, grounded, grouped, and presented to the reasoning model. 
By executing the full reasoning pipeline, collecting failed rollouts, and reflecting on representation-level defects such as missing constraints, ambiguous targets, distracting relations, and unmapped angle labels, GeoReform turns downstream reasoning failures into targeted policy mutations. This process encourages formalizations that are complete enough to support the solution while remaining concise enough to avoid distracting the reasoning.

Our contributions can be summarized as follows:
\begin{itemize}
  \item We show that extracting geometric facts alone is insufficient for reliable geometry reasoning. Effective formalization must therefore organize these facts into selective and unambiguous representations.
  
    \item We introduce GeoReform, a reflective formalization evolution framework that optimizes formalization policies through genetic search and reflective formalization mutation.

    \item We show that evolving the formalization policy mitigates these failures and substantially improves GPS performance, raising accuracy from 42.0\% to 56.0\% on Geometry3K.
\end{itemize}
\section{Related Work}

\textbf{Geometry Problem Solving}
Geometry problem solving (GPS) requires accurate diagram understanding and rigorous mathematical reasoning. Early systems such as Inter-GPS parse textual and visual inputs into formal representations for symbolic reasoning with theorem knowledge~\citep{lu2021intergps,zhang2023pgpsnet,peng2023geodrl}. Neural approaches explore intermediate representations such as geometric primitives, solution programs, and mathematical expressions~\citep{zhang2022plane,chen2021geoqa,chen2022unigeo}. AlphaGeometry further combines neural search with symbolic deduction for geometry theorem proving~\citep{trinh2024solving}. Recent MLLM methods improve geometric reasoning through geometry instruction data and formalized vision-language pretraining, as exemplified by G-LLaVA and GeoX~\citep{gao2025gllava,xia2025geox}. AutoGPS, GOLD, and DFE-GPS explicitly incorporate structured diagram information into problem solving~\citep{ping2025autogps,zhang2024gold,zhang2024diagram}. These approaches highlight the value of geometric structure, but errors in grounding, relation selection, or target specification can affect downstream reasoning. GeoReform studies how formalized information should be selected, grounded, and organized to support problem solving.

\paragraph{Reflective Prompt Evolution}
Prompt evolution iteratively searches, evaluates, and refines prompts. CoT demonstrates how reasoning instructions affect model behavior~\citep{wei2022chain}, while APE, ProTeGi, and OPRO optimize instructions through candidate search, textual feedback, or LLM-based optimization~\citep{zhou2023large,pryzant2023automatic,yang2024large}. EvoPrompt and PromptBreeder evolve prompt populations through mutation and selection~\citep{guo2024connecting,fernando2023promptbreeder}. Self-Refine, Reflexion, PromptAgent, and GEPA further use model errors or execution feedback to guide improvement~\citep{madaan2023selfrefine,shinn2023reflexion,wang2024promptagent,agrawal2026gepa}. Such feedback can inform both prompt revisions and reusable rules for addressing failures. Existing methods generally optimize task instructions, exemplars, or reasoning prompts. GeoReform applies evolution to the intermediate representation within the GPS pipeline: it keeps the downstream reasoning prompt fixed and evolves the formalization policy governing which geometric entities, relations, constraints, and targets are extracted and how they are grounded, organized, and presented. This allows reasoning feedback to improve the geometric information supplied to the model.
\section{Method}
\subsection{Method Overview}
% 画图突出关系三大块

% 颜色 雷达图 
% 小红书nature模板 /2 3个表 /4 5个图 case study

% 字和框离得太近了
% 表达的东西 字体太小 
% 雷达图 有图例说明 2B和4B

\begin{figure*}[t]
    \centering
    \includegraphics[width=\textwidth]{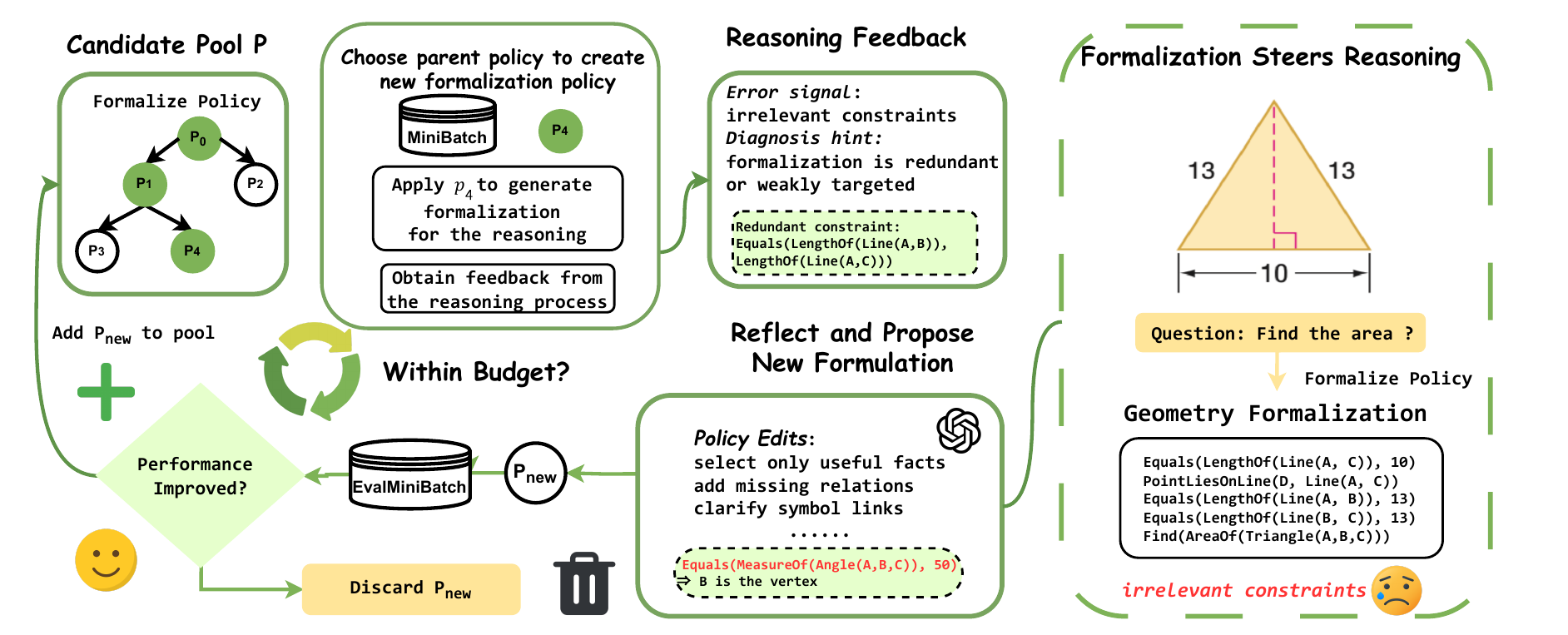}
    \vspace{-5mm}
   \caption{\small GeoReform optimizes the policy that turns a geometry problem into a formal representation. In each iteration, GeoReform reflects on failed reasoning cases from an existing formalization policy and proposes a new policy. The new policy is first tested on a minibatch, and if improved, evaluated on a validation set. Successful policies are added to the candidate pool, guiding formalization toward clearer and more useful geometric representations.
}\label{fig:mathpipeline}
    \vspace{-5mm}
\end{figure*}

To address the representation bottleneck in multimodal geometry reasoning, we introduce GeoReform, a reflective formalization evolution framework that optimizes how geometric entities, relations, constraints, and targets are organized for downstream reasoning. The overall pipeline is illustrated in Figure~\ref{fig:mathpipeline}.

\paragraph{Formalization Policy Design.}
Prior GPS systems commonly rely on intermediate formal representations to bridge diagram understanding and downstream solving~\citep{lu2021intergps, chen2021geoqa, chen2022unigeo, zhang2023pgpsnet, xia2025geox}. 
While such representations can expose useful geometric structures, they may also introduce redundant relations or ambiguously grounded constraints that interfere with reasoning. 
As detailed in Sec.~\ref{sec:interface}, GeoReform makes the formalization policy explicit, defining how to select, ground, group, and present geometric entities, relations, constraints, and targets for the reasoning model. This explicit policy view allows GeoReform to treat formalization as an optimizable interface rather than a fixed parser output.

\paragraph{Reflective Formalization Evolution.}
To better exploit formalization, we introduce reflective formalization evolution in Sec.~\ref{sec:evolution}. 
GeoReform starts from an initial formalization policy, evaluates its downstream reasoning behavior, and uses failure cases to identify weaknesses in the current representation. 
These weaknesses are then converted into targeted policy updates, allowing GeoReform to iteratively refine how geometric information is selected, grounded, grouped, and presented.

During inference, the evolved formalization policy is fixed and applied to each test problem to generate a structured representation of the relevant geometric information. 
The downstream reasoning model then conditions on both the original problem and this representation to finish reasoning and predict the final answer.

\subsection{Formalization Policy Design}
\label{sec:interface}

To make geometric conditions explicit for downstream reasoning, we formulate formalization as an intermediate interface between multimodal problem understanding and reasoning. 
A formalization policy $p$ defines how geometric information should be extracted, organized, and presented.
Given a diagram $I$ and a problem statement $T$, the structure generator $G$ follows $p$ to produce a formalized representation $S_p = G(I, T; p)$.
Inspired by prior GPS systems that use formal geometric structures to support downstream solving~\citep{lu2021intergps, chen2021geoqa, chen2022unigeo, zhang2023pgpsnet, xia2025geox}, we first define a geometry language space $\Omega$ following \citet{lu2021intergps}. 
This space consists of typed literals, where each literal applies a predicate to a set of arguments and describes a geometric entity, relation, numerical constraint, or target query. 
For instance, a formalized representation may contain literals such as \texttt{Triangle(A,B,C)}, \texttt{Parallel(Line(A,B),Line(C,D))}and \texttt{Equals(Angle(A,B,C),54)}.

Based on this language space, we define an initial formalization policy $p_0$ as the starting point for the reflective evolution process described in Sec.~\ref{sec:evolution}. 
This policy specifies how the structure generator should use the typed literals to describe a geometry problem, and the literals in $S_{p_0}$ provide an explicit structured description of the original problem. The policy also controls the output form. It must not return explanations, derivations, or extra text.
This keeps the formalization short and reduces noise for the reasoning.
For any formalization policy $p$, the generated representation $S_p$ is then provided to the reasoning model together with the original problem:
$
\hat{a} = \mathrm{Solver}(I, T, C, S_p),
$
where $C$ denotes the candidate answer choices when available. During evolution, GeoReform updates only the policy. The best policy is fixed and used for all test problems.

\subsection{Reflective Formalization Evolution}
\label{sec:evolution}

\subsubsection{Genetic Optimization Loop}
\begin{table*}[t]
\centering
\caption{\small Performance comparison of formalization evolution on Geometry3K. All values are accuracy (\%). Improvements over the initial policy are shown in parentheses.}
\label{tab:main_results}

\resizebox{0.8\textwidth}{!}{%
\fontsize{9pt}{10pt}\selectfont%
\setlength{\tabcolsep}{8pt}%
\renewcommand{\arraystretch}{1.15}%
\setlength{\aboverulesep}{0pt}%
\setlength{\belowrulesep}{0pt}%
\newcommand{\gain}[1]{{\fontsize{8.0pt}{8.2pt}\selectfont\textcolor{gray}{$(+#1)$}}}%
\newcommand{\res}[2]{#1\gain{#2}}%
\begin{tabular}{ll|cc|cc|cc}
\toprule
\rowcolor{gray!15}
Backbone & Scale
& \multicolumn{2}{c|}{MLLM}
& \multicolumn{2}{c|}{+ Initial Policy}
& \multicolumn{2}{c}{+ GeoReform} \\
\cline{3-8}
\rowcolor{gray!15}
&
& Choice & Completion
& Choice & Completion
& Choice & Completion \\
\midrule
\multirow{4}{*}{Qwen3VL}
& 2B  & 33.5 & 29.0 & 42.0 & 37.0 & \res{56.0}{14.0} & \res{53.5}{16.5} \\
& 4B  & 55.0 & 51.5 & 69.5 & 67.0 & \res{78.0}{8.5}  & \res{73.0}{6.0} \\
& 8B  & 69.0 & 54.0 & 74.0 & 68.5 & \res{79.0}{5.0}  & \res{74.0}{5.5} \\
& 32B & 76.5 & 70.0 & 82.0 & 74.5 & \res{88.0}{6.0}  & \res{80.0}{5.5} \\
\midrule
\multirow{5}{*}{InternVL3.5}
& 1B  & 41.5 & 28.0 & 43.0 & 27.5 & \res{57.5}{14.5} & \res{40.5}{13.0} \\
& 2B  & 50.5 & 29.5 & 54.0 & 36.0 & \res{68.0}{14.0} & \res{53.5}{17.5} \\
& 4B  & 53.0 & 37.0 & 65.0 & 48.5 & \res{73.0}{8.0}  & \res{57.0}{8.5} \\
& 8B  & 61.0 & 43.5 & 73.5 & 60.0 & \res{77.5}{4.0}  & \res{66.5}{6.5} \\
& 14B & 65.5 & 48.5 & 74.0 & 59.0 & \res{78.5}{4.5}  & \res{67.5}{8.5} \\
\bottomrule
\end{tabular}
}
\vspace{-2mm}
\end{table*}
\label{sec:genetic_loop}

Given a geometry reasoning system $\Phi$ consisting of a structure generator and a downstream reasoning model, GeoReform aims to identify a formalization policy that maximizes downstream solving performance. 
GeoReform starts from the initial policy $p_0$ described in Sec.~\ref{sec:interface} and maintains a candidate pool $\mathcal{P}$ of formalization policies. 
During evolution, we also maintain a best policy candidate according to performance on an evaluation minibatch. 
GeoReform then enters an optimization loop, repeatedly proposing new candidate policies until the evaluation budget is exhausted. The overall reflective formalization evolution procedure is summarized in Algorithm~\ref{alg:reform}.

At each iteration, GeoReform proceeds in four steps:
(i) it selects a parent policy from the candidate pool, biased toward the current best policy while still allowing random exploration;
(ii) it runs the full reasoning pipeline on a training minibatch and collects failed cases, including the problem statement, diagram, extracted geometric relations, reasoning process, prediction, and evaluation feedback;
(iii) it sends these failed cases to the reflection model, which produces a child policy by revising the current formalization policy;
and (iv) it evaluates the child policy on the same training minibatch as an efficient filter. 
If the child policy outperforms its parent, it is added to the candidate pool and further evaluated on an evaluation minibatch. 
If its evaluation accuracy is no lower than the best historical evaluation accuracy, it becomes the new best policy candidate.

The training minibatch is used for failure collection and efficient filtering, while the evaluation minibatch is used to track the best policy candidate. 
The final test set is held out and is not used during evolution.

\subsubsection{Reflective Formalization Mutation}
\label{sec:reflective_mutation}

Rollout traces reveal how a formalization shapes downstream reasoning, not just whether the final answer is correct. 
This is crucial because poor formalizations fail in opposite ways: verbose formalizations distract the reasoning model with irrelevant relations, while overly conservative formalizations omit necessary constraints. 
By pairing rollout traces with final outcomes, GeoReform turns single answer failures into representation diagnoses. 
The reflection model compares the generated formalization with the reasoning trajectory and attributes failures to concrete policy defects. 
Reflective mutation then converts these diagnoses into targeted policy updates, revising the formalization policy to preserve necessary constraints, remove distractions, clarify targets, and better organize geometric information.

\begin{algorithm}[!b]
\caption{\small Reflective Formalization Evolution}
 % \hua{This algorithm is never referred}
\label{alg:reform}
\footnotesize
\LinesNotNumbered

\KwIn{initial policy $p_0$, training set $D_{\mathrm{train}}$, evaluation minibatch $B_{\mathrm{eval}}$, selection threshold $\tau$, evolution budget $T$}
\KwOut{best policy $p^*$}

$\mathcal{P}\leftarrow\{p_0\}$, $p^*\leftarrow p_0$, $s^*\leftarrow \mathrm{score}(p^*,B_{\mathrm{eval}})$\;

\For{$t = 1, \ldots, T$}{
    Sample $u \sim \mathrm{Uniform}(0,1)$ and set 
    $p_t \leftarrow p^*$ if $u<\tau$, otherwise $p_t \sim \mathrm{Uniform}(\mathcal{P})$\tcp*{select parent}

    Sample training minibatch $B_t \subset D_{\mathrm{train}}$\;
    Run $\Phi(p_t)$ on $B_t$ and collect rollouts $\mathcal{R}_t$\;
    Extract failed cases $\mathcal{F}_t \subset \mathcal{R}_t$\tcp*{bad cases}

    Construct reflection inputs from $\mathcal{F}_t$ and generate child policy 
    $p_t' = R(\mathcal{F}_t,p_t)$\tcp*{mutation}

    \If{$\mathrm{score}(p_t',B_t) > \mathrm{score}(p_t,B_t)$}{
        Add $p_t'$ to $\mathcal{P}$ and evaluate $p_t'$ on $B_{\mathrm{eval}}$\tcp*{filter}
        \If{$\mathrm{score}(p_t',B_{\mathrm{eval}}) \ge s^*$}{
            $p^* \leftarrow p_t'$, 
            $s^* \leftarrow \mathrm{score}(p_t',B_{\mathrm{eval}})$\tcp*{update best}
        }
    }
}
\Return{$p^*$}\;
\end{algorithm}

Given a parent policy selected by the genetic optimization loop, GeoReform executes it on a minibatch and collects failed rollouts. 
For each failure, the reflection model compares the generated formalization with the reasoning trajectory and evaluation feedback. 
This comparison exposes the likely policy defect: wrong quantity computation may suggest weak target binding, angle confusion may suggest missing angle mappings, irrelevant derivations may suggest redundant or poorly grouped relations, and incomplete reasoning may suggest missing constraints. 
Thus, reflection converts final answer errors into actionable feedback on the formalization policy.

Mutation updates only the formalization policy; the structure generator, reasoning model, and reasoning prompt are kept fixed. 
The update is a general instruction revision rather than a correction to a specific example. 
It may require explicit targets, clearer grouping of entities and constraints, useful properties of named figures, or restraints against unsupported diagram assumptions. 
The child policy is then returned to the genetic loop for gated evaluation and pool update.

\section{Experiment}
\label{sec:exp}
\begin{table*}[t]
\centering
\caption{\small Performance comparison of different policies on Geometry3K using InternVL3.5. All values are accuracy (\%). Improvements of GeoReform over the best competing method are shown in parentheses.}
\label{tab:internvl35_results}

\resizebox{0.8\textwidth}{!}{%
\fontsize{9pt}{10pt}\selectfont%
\setlength{\tabcolsep}{6pt}%
\renewcommand{\arraystretch}{1.15}%
\setlength{\aboverulesep}{0pt}%
\setlength{\belowrulesep}{0pt}%

\newcommand{\gain}[1]{{\fontsize{8.0pt}{8.2pt}\selectfont\textcolor{gray}{$(+#1)$}}}%
\newcommand{\res}[2]{#1\gain{#2}}%

\begin{tabular}{ll|c|ccc|c|c}
\toprule
\rowcolor{gray!15}
Setting & Scale
& MLLM
& \multicolumn{3}{c|}{AutoGPS}
& InterGPS
& GeoReform \\

\cline{4-6}

\rowcolor{gray!15}
&
&
& GPT4o & InternVL3 & Qwen2.5
&
& \\
\midrule

\rowcolor{gray!8}
\multicolumn{8}{c}{\textbf{Choice}} \\
\midrule
\multirow{5}{*}{InternVL3.5}
& 1B  & 41.5 & 44.5 & 38.5 & 40.5 & 41.0 & \res{57.5}{13.0} \\
& 2B  & 50.5 & 50.5 & 49.0 & 49.5 & 52.5 & \res{68.0}{15.5} \\
& 4B  & 53.0 & 56.0 & 56.5 & 62.0 & 62.0 & \res{73.0}{11.0} \\
& 8B  & 61.0 & 60.0 & 59.5 & 62.0 & 63.5 & \res{77.5}{14.0} \\
& 14B & 65.5 & 68.5 & 66.5 & 67.5 & 66.0 & \res{78.5}{10.0} \\

\midrule
\rowcolor{gray!8}
\multicolumn{8}{c}{\textbf{Completion}} \\
\midrule
\multirow{5}{*}{InternVL3.5}
& 1B  & 28.0 & 32.5 & 30.0 & 28.0 & 25.0 & \res{40.5}{8.0} \\
& 2B  & 29.5 & 28.5 & 28.5 & 31.5 & 32.5 & \res{53.5}{21.0} \\
& 4B  & 37.0 & 43.5 & 45.0 & 42.5 & 44.5 & \res{57.0}{12.0} \\
& 8B  & 43.5 & 48.5 & 50.5 & 49.0 & 58.5 & \res{66.5}{8.0} \\
& 14B & 48.5 & 54.5 & 56.0 & 54.0 & 60.0 & \res{67.5}{7.5} \\

\bottomrule
\end{tabular}
}
\vspace{-2mm}
\end{table*}

\subsection{Experiment Setup}
\label{sec:exp_setup}

% \vspace{1mm}
\textbf{Datasets.}
Experiments were conducted on three specialized GPS benchmarks: Geometry3K~\citep{lu2021intergps}, GeoQA~\citep{chen2021geoqa}, and PGPS9K~\citep{zhang2023}. 
Geometry3K consists of plane geometry problems with diagrams, textual questions, answer choices, and formal annotations, and is designed to support geometry parsing and symbolic reasoning. 
GeoQA focuses on geometry question answering, where many geometric conditions are described in the problem text and accompanied by diagrams; we use its English version to maintain linguistic consistency with the other datasets. Since GeoQA lacks official formal geometric annotations, we use GPT-5 to extract structured geometric information from each problem (text and diagram). PGPS9K~\citep{zhang2023} contains 9,022 plane geometry problems with fine-grained diagram annotations and interpretable solution programs.

\vspace{1mm}
\paragraph{Metrics.}
We follow the evaluation protocol used in prior work on Geometry3K~\citep{lu2021intergps}. 
Specifically, we evaluate model performance using Completion and Choice: 
In \textit{Completion}, models directly generate the final answer without being provided candidate options, while in \textit{Choice}, the model is provided with possible options. 
These metrics jointly assess the model's ability to perform direct geometric reasoning, option-based answer selection, and ranked prediction. 
% \hua{Completion and Choice seems to be two modes, rather than metrics - metrics is about how to calculate values, like Accuracy}

\vspace{1mm}
\begin{table}[!b]
\centering
\caption{\small Effect of explicit geometric structure on GeoQA and PGPS9K. Most geometric information in GeoQA is provided in the problem statement; PGPS9K uses ground-truth geometric annotations. All values are accuracy (\%).}
\label{tab:geometry_structure}

\resizebox{0.99\linewidth}{!}{%
\fontsize{6.3pt}{6.6pt}\selectfont
\setlength{\tabcolsep}{3.6pt}
\renewcommand{\arraystretch}{1.08}
\setlength{\aboverulesep}{0pt}
\setlength{\belowrulesep}{0pt}

\newcommand{\gtgain}[1]{{\fontsize{5.8pt}{6.0pt}\selectfont\textcolor{gray}{(#1)}}}
\newcommand{\gtres}[2]{#1\gtgain{#2}}

\begin{tabular}{ll|cc}
\toprule
\rowcolor{gray!15}
Setting & Scale
& \multicolumn{2}{c}{Metric} \\
\cline{3-4}

\rowcolor{gray!15}
&
& Baseline & + Geometry Structure \\
\midrule

\multicolumn{4}{c}{GeoQA -- Choice} \\
\midrule

\multirow{4}{*}{Qwen3VL}
& 2B  & 49.0 & \gtres{56.5}{+7.5} \\
& 4B  & 78.5 & \gtres{81.5}{+3.0} \\
& 8B  & 81.0 & \gtres{80.0}{-1.0} \\
& 32B & 92.0 & \gtres{90.0}{-2.0} \\

\midrule

\multicolumn{4}{c}{GeoQA -- Completion} \\
\midrule

\multirow{4}{*}{InternVL3.5}
& 1B & 23.5 & \gtres{24.0}{+0.5} \\
& 2B & 32.0 & \gtres{34.0}{+2.0} \\
& 4B & 51.0 & \gtres{55.5}{+4.5} \\
& 8B & 58.5 & \gtres{64.0}{+5.5} \\

\midrule

\multicolumn{4}{c}{PGPS9K -- Choice} \\
\midrule

\multirow{4}{*}{Qwen3VL}
& 2B  & 32.0 & \gtres{37.5}{+5.5} \\
& 4B  & 54.5 & \gtres{76.0}{+21.5} \\
& 8B  & 66.5 & \gtres{76.5}{+10.0} \\
& 32B & 76.5 & \gtres{87.0}{+10.5} \\

\bottomrule
\end{tabular}
}
\end{table}

\paragraph{Baselines.}
We compare GeoReform with three baselines. 
(i) \textit{Initial Policy} uses the same structure generator and downstream reasoning prompt as GeoReform, but keeps the original formalization policy fixed. This baseline isolates the effect of reflective formalization evolution. 
(ii) InterGPS~\citep{lu2021intergps} parses text and diagrams into formal geometric relations using an OCR based pipeline. 
(iii) AutoGPS~\citep{ping2025autogps} produces multimodal geometric formalizations and combines them with deductive reasoning. Although these baselines obtain geometric structures from different sources, we feed all formalization inputs to the same downstream reasoning model with the same reasoning prompt, decoding setting, and evaluation protocol, treating them as alternative formalization inputs under a unified evaluation setting.

\vspace{1mm}
\begin{figure*}[t]
    \centering
    \includegraphics[width=\textwidth]{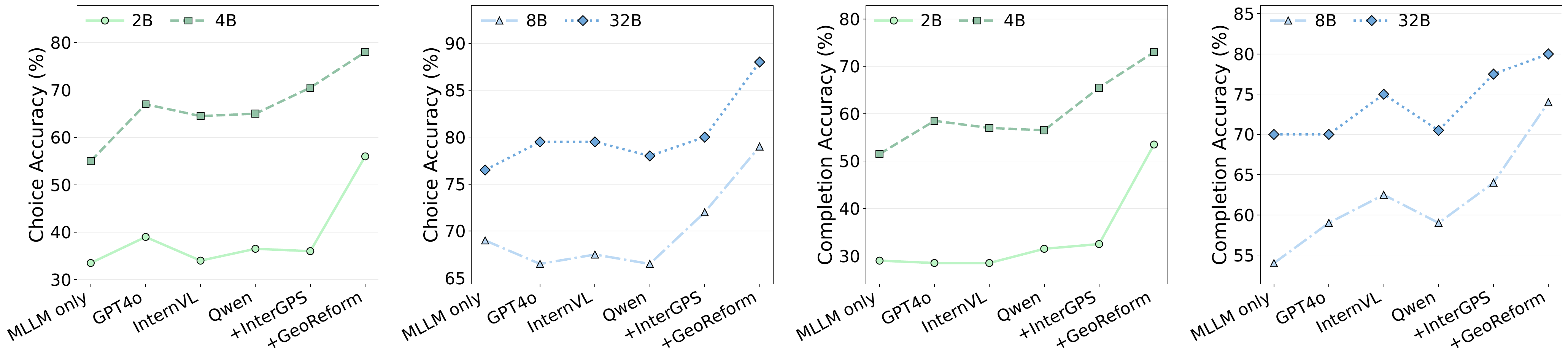}
\vspace{-5mm}
\caption{\small Comparison of formalization methods on Geometry3K using Qwen3VL reasoning models. \textit{MLLM only} uses no additional formalization; \textit{GPT4o}, \textit{InternVL}, and \textit{Qwen} indicate the models providing formalizations for AutoGPS.}
    \label{fig:autogps2}
\vspace{-4mm}
\end{figure*}
\begin{figure*}[t]
    \centering
    \includegraphics[width=\textwidth]{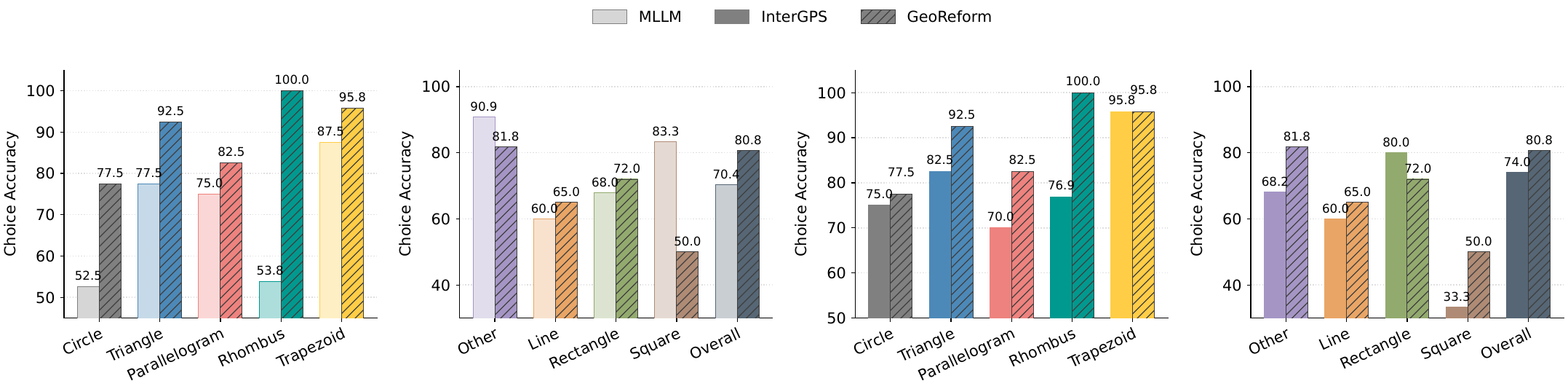}
    \vspace{-5mm}
    \caption{\small
   Choice accuracy by geometry problem type on Geometry3K on Qwen3VL-8B model. GeoReform achieves the strongest gains on Rhombus and Parallelogram, where solving depends on global shape relations rather than only local marks.
    }
    \label{fig:type_accuracy}
\end{figure*}

\paragraph{Implementation Details.}
We use GPT-5 as the structure generation model to provide a strong pool of geometrically valid facts, allowing us to focus on optimizing the formalization policy.
Reflective evolution runs for at most 20 update iterations.
At each iteration, we randomly sample 32 training problems for failure collection and policy filtering, using a fixed random seed of 42 across all experiments.
The reflection input contains at most 16 failed cases from the sampled minibatch, including the original problem, generated formalization, downstream reasoning trace, model prediction, and gold answer.
We use a fixed validation minibatch of 50 problems to select the best evolved policy.
The final test examples are held out and never used during evolution.
Unless otherwise specified, policy selection is based on choice accuracy.
For downstream reasoning, we evaluate models from the Qwen3VL~\citep{bai2025qwen3vl} and InternVL3.5~\citep{wang2025internvl35} series, using the same reasoning prompt across all policies.
We serve models with vLLM~\citep{kwon2023efficient} and set the maximum generation length to 4096 tokens.
All experiments are conducted on NVIDIA A6000 and NVIDIA ProA6000 GPUs.

\subsection{Experiment Results}
\begin{figure*}[t]
    \centering
    \includegraphics[width=\textwidth]{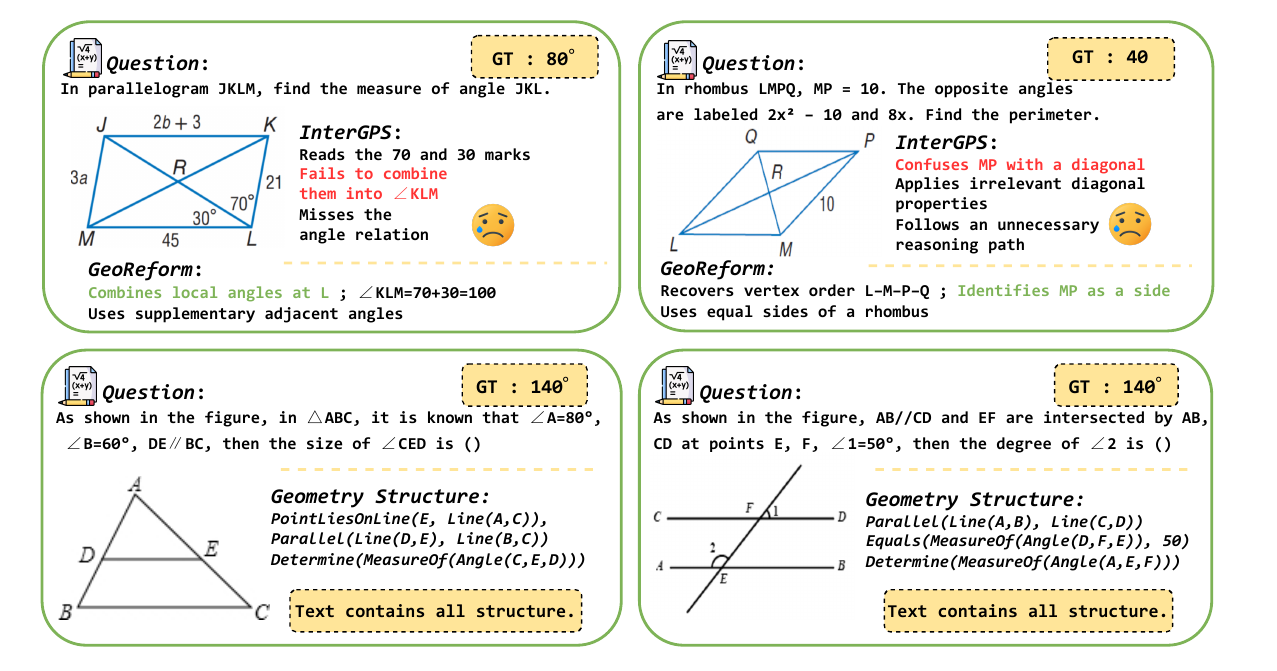}
\vspace{-8mm}
\caption{\small
Case studies of geometric formalization. 
Top: GeoReform improves over InterGPS by linking local marks to global shape relations. Bottom: GeoQA cases show that all required structural information is contained in the question text. Even in this setting, making the structure explicit helps organize the stated facts for more reliable reasoning.
}
    \label{fig:case_analysis}
\end{figure*}

\paragraph{Reflective Evolution Improves Formalization Quality.}
\label{sec:evolution_results}
As shown in Table~\ref{tab:main_results}, GeoReform improves over the initial policy described in Sec.~\ref{sec:interface} across all evaluated backbones, scales, and settings, with gains of 4.0\%--17.5\%.
On Qwen3VL-2B, choice accuracy increases from 42.0\% to 56.0\%, while completion accuracy rises from 37.0\% to 53.5\%.
Importantly, the initial policy does not always help: it slightly reduces InternVL3.5-1B completion accuracy from 28.0\% to 27.5\%, whereas GeoReform raises it to 40.5\%.
These results highlight the value of using reasoning failures to refine formalization.
A representation may expose geometric facts yet still leave the model uncertain about which constraints matter or how symbols map to the diagram.
Policy evolution addresses this gap by improving how the available facts are selected, grounded, and organized.

\paragraph{Sensitivity to Representation Quality.}
The largest gains generally occur on 1B--2B models, reaching 17.5\% on InternVL3.5-2B completion.
This pattern suggests that smaller models have more difficulty compensating for unclear or distracting representations.
Making relevant constraints explicit can reduce the need to recover missing connections or resolve ambiguous symbols during reasoning.
The gains are not strictly monotonic with model size, however, and the largest evaluated models still improve by 4.5\%--8.5\%.
Thus, stronger model capacity reduces some dependence on representation quality but does not remove the benefit of refining the formalization.

\paragraph{Comparison with Explicit Formalization.}
Table~\ref{tab:internvl35_results} and Figure~\ref{fig:autogps2} compare GeoReform with InterGPS~\citep{lu2021intergps} and AutoGPS~\citep{ping2025autogps}.
GeoReform performs best across all evaluated settings.
On InternVL3.5, it exceeds the best competing method by 10.0\%--15.5\% in choice and 7.5\%--21.0\% in completion.
Some competing formalizations nevertheless perform below the original MLLM baseline, showing that an additional structured input can also interfere with reasoning.
The comparison suggests that extracting or aligning geometric relations is insufficient by itself: the resulting representation must make the relevant constraints easy to identify and use.
GeoReform directly optimizes this downstream usefulness through feedback from failed solutions.

\subsubsection{Ablation Study}
\label{sec:ablation}

\paragraph{Effect of Explicit Geometric Structure.}
Table~\ref{tab:geometry_structure} shows that explicit geometric structure improves performance on both GeoQA and PGPS9K.
Even on GeoQA, where most geometric conditions are already stated in the problem text (Figure~\ref{fig:case_analysis}), adding structure improves six of eight settings, with gains of up to 7.5\%.
On PGPS9K, ground-truth geometric annotations improve choice accuracy across all evaluated Qwen3VL scales by 5.5\%--21.5\%.
These gains suggest that explicit structure helps not only by supplying information from diagrams, but also by making available geometric conditions easier to use in reasoning.
The small drops in two GeoQA settings further indicate that the benefit depends on how well the added structure complements the original input.

\subsection{Case Analysis}

\paragraph{GeoReform Connects Marks to Shape Relations.}

GeoReform improves because it produces geometric relations that are more useful for the target question, capturing local marks such as points, lines, angles, and lengths while connecting them to the full shape. In contrast, InterGPS improves less because its output remains more local. Although it can detect many useful facts, it often leaves them as separate pieces. As a result, its representation may show that a rhombus or a parallelogram is present, but still fail to specify how each mark should be used. For example, it may confuse a side with a diagonal, or miss that two marked angles form one full angle. This pattern is clear across geometry types. As illustrated in Figure~\ref{fig:type_accuracy}, GeoReform gains the most on Rhombus and Parallelogram. On Rhombus, GeoReform reaches $100.0\%$ accuracy, while InterGPS reaches only $76.9\%$. On Parallelogram, GeoReform reaches $82.5\%$, compared with $70.0\%$ for InterGPS. These types benefit more because their answers depend strongly on global shape relations. Rhombus problems require relations such as equal sides, diagonal bisection, and the link between half diagonals and full diagonals. Parallelogram problems often require opposite sides, adjacent angles, and angle composition. GeoReform recovers these relations more reliably, leading to more accurate answers. 
The evolution trace of the selected policy is reported in Appendix~\ref{app:evolution_trace}. We provide all prompts used by GeoReform, including the initial formalization prompt, reflection prompt, downstream reasoning prompt, and final evolved prompt, in Appendix~\ref{app:prompt_policy_details}.

\section{Conclusion}

In this paper, we proposed GeoReform, a reflective formalization evolution framework for multimodal Geometry Problem Solving (GPS). GeoReform shows that effective formalization is not merely about extracting geometric facts, but about organizing them into representations that better support downstream reasoning. By optimizing the formalization policy through feedback from failed rollouts, GeoReform refines how geometric entities, relations, constraints, and targets are selected, grounded, grouped, and presented. Extensive experiments and analyses on multiple geometry reasoning benchmarks demonstrate that GeoReform consistently improves over the initial policy and achieves strong performance across different reasoning models.

\section*{Limitations}
\label{sec:limitation}

Although GeoReform improves multimodal geometry problem solving by evolving the formalization policy, it still has several limitations. First, our experiments mainly focus on plane geometry benchmarks, which do not fully represent broader mathematical reasoning scenarios, such as solid geometry, algebraic diagrams, physics problems, or proof-oriented geometry tasks. Future work should examine whether reflective formalization evolution can generalize to these more diverse settings. Second, while GeoReform improves how geometric information is selected, grounded, grouped, and presented, it does not completely eliminate errors from diagram understanding or structure generation. If the underlying model misses a visual mark, misidentifies a geometric entity, or grounds an angle or segment incorrectly, the resulting formalization may still mislead downstream reasoning. Incorporating stronger visual parsers, symbolic consistency checks, or theorem-based verification could further improve the reliability of the generated representations. Third, the current evaluation mainly reports accuracy without full statistical significance analysis across repeated runs, since repeated large-scale multimodal inference across many backbones is computationally expensive.

\bibliography{references}

\onecolumn
\appendix
\section{Appendix}
\subsection{Evolution Trace}
\label{app:evolution_trace}

\begin{table*}[!h]
\centering
\caption{\small Trace of reflective formalization evolution. We show only policies on the final best path; other rejected or neutral candidates are omitted.}
\label{tab:evolution_trace}
% \vspace{-2mm}
\fontsize{8.0pt}{8.3pt}\selectfont
\setlength{\tabcolsep}{5.0pt}
\renewcommand{\arraystretch}{1.35}
\setlength{\aboverulesep}{0pt}
\setlength{\belowrulesep}{0pt}

\newcommand{\evgain}[1]{{\fontsize{7.0pt}{7.2pt}\selectfont\textcolor{gray}{(#1)}}}
\newcommand{\evres}[2]{#1\evgain{#2}}
\newcommand{\bestres}[2]{\textbf{#1}\evgain{\textbf{#2}}}

\begin{tabularx}{\textwidth}{c c c X c c}
\toprule
\rowcolor{gray!15}
Path & Policy & Parent & Policy Change & Train & Val \\
\midrule

0 & $P_0$ & -- 
& Initial conservative parser in the style of InterGPS. 
& 68.75 & 75.00 \\

1 & $P_3$ & $P_0$ 
& General reflective revision with stronger fitness on the train batch. 
& 81.25 & \evres{79.17}{+4.17} \\

2 & $P_{12}$ & $P_3$ 
& Explicit mapping for entities, angles, lines, transversals, and targets. 
& 68.75 & \evres{83.33}{+8.33} \\

\rowcolor{green!8}
3 & $P_{20}$ & $P_{12}$ 
& Named figure policy with restrained expansion of useful properties. 
& 75.00 & \bestres{87.50}{+12.50} \\

\bottomrule
\end{tabularx}
\vspace{-0.6em}
\end{table*}

Table~\ref{tab:evolution_trace} shows the final policy lineage selected by reflective evolution, where Path denotes lineage depth rather than search order. Starting from the initial policy $P_0$, the search first improves general formalization quality with $P_3$, and then shifts toward explicit grounding of entities, angles, lines, and targets in $P_{12}$. Although $P_{12}$ has lower training accuracy than $P_3$, its validation accuracy is higher, suggesting that better formalization is not simply obtained by fitting the training batch. This highlights the importance of using the validation set to select policies that generalize beyond the sampled failures. The final policy $P_{20}$ further adds named figure information while keeping property expansion restrained, achieving the best validation accuracy of $87.50\%$.

\subsection{Prompts and Policy Details}
\label{app:prompt_policy_details}

This appendix reports the prompts and policies used by GeoReform. The initial formalization policy defines the starting representation interface, the reflection prompt specifies how failed rollouts are converted into policy mutations, the reasoning prompt is kept fixed during evolution, and the final evolved policy is used for all test examples.

\subsubsection{Initial Formalization Policy}
\label{app:initial_policy}

\begin{figure*}[!t]
    \centering
    \includegraphics[width=\textwidth]{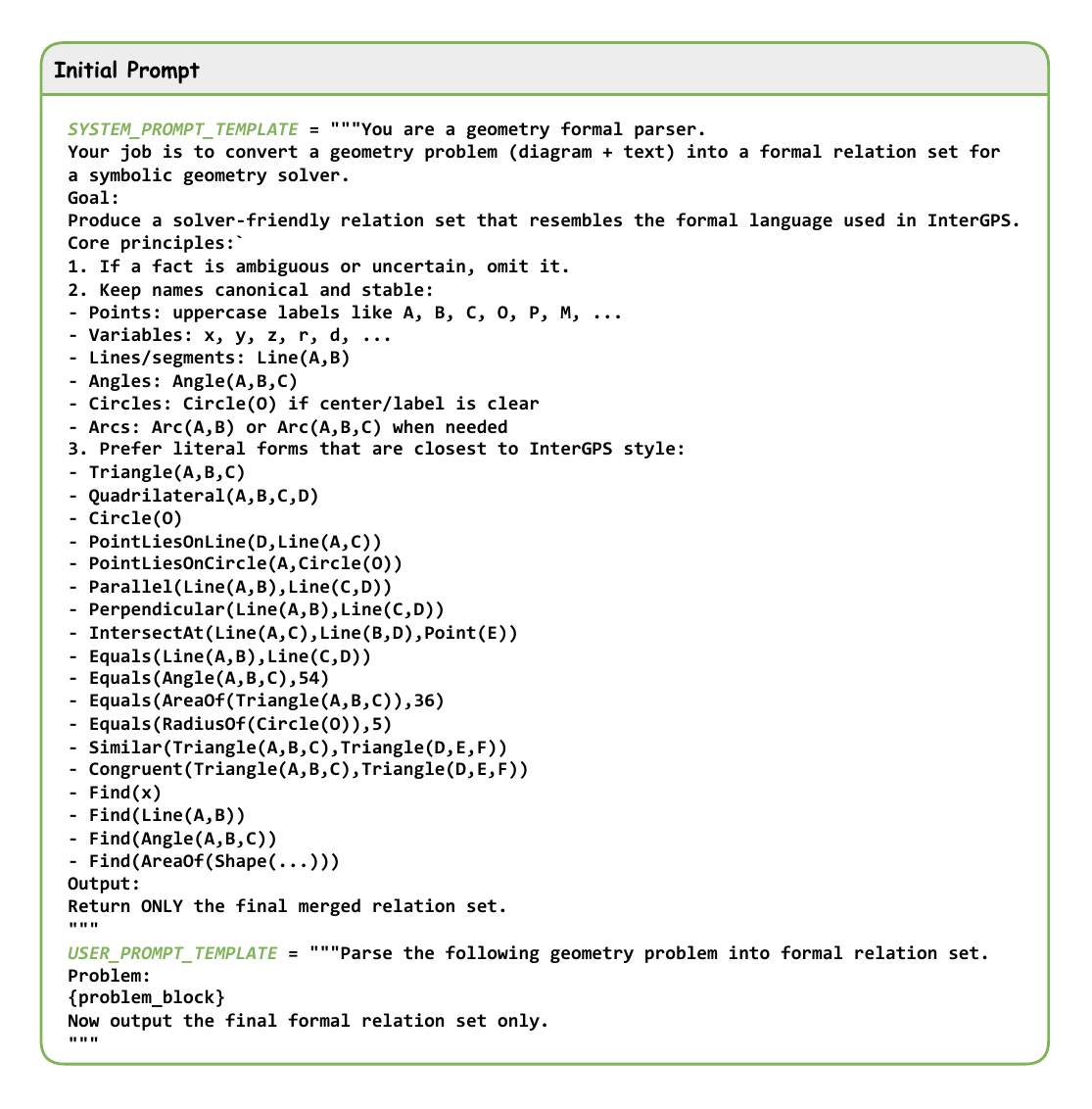}
    \vspace{-4mm}
    \caption{\small Initial formalization policy used by the structure generator. The policy specifies how geometric entities, incidence relations, polygons, angle relations, length constraints, and targets are extracted and organized before downstream reasoning.}
    \label{fig:appendix_initial_policy_prompt}
    \vspace{-2mm}
\end{figure*}

\subsubsection{Reflection Prompt for Policy Mutation}
\label{app:reflection_prompt}

\begin{figure*}[!t]
    \centering
    \includegraphics[width=\textwidth]{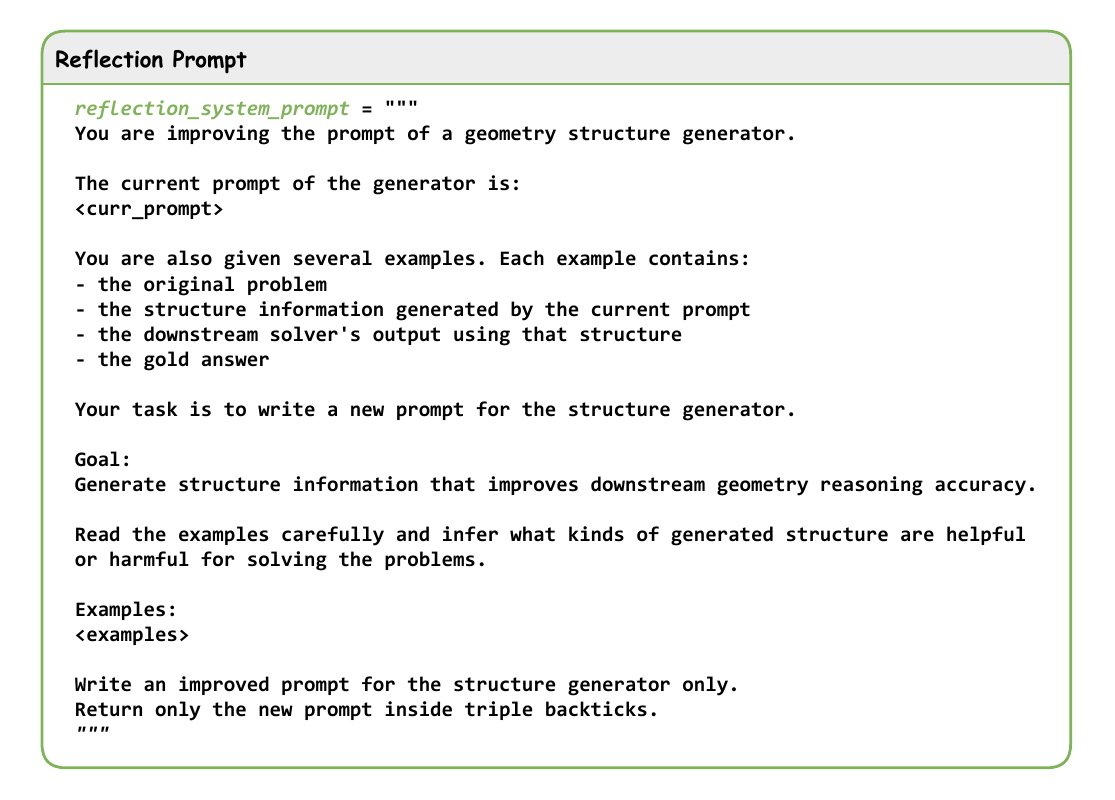}
    \vspace{-4mm}
    \caption{\small Reflection prompt used to diagnose failed rollouts and generate a revised formalization policy. }
    \label{fig:appendix_reflection_prompt}
    \vspace{-2mm}
\end{figure*}

\subsubsection{Downstream Reasoning Prompt}
\label{app:reasoning_prompt}

\begin{figure*}[!t]
    \centering
    \includegraphics[width=\textwidth]{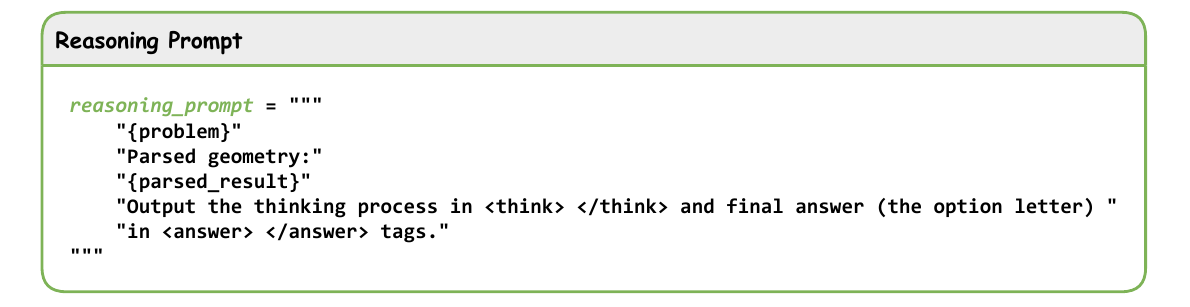}
    \vspace{-4mm}
    \caption{\small Downstream reasoning prompt used by the reasoning model.}
    \label{fig:appendix_reasoning_prompt}
    \vspace{-2mm}
\end{figure*}

\subsubsection{Final Evolved Formalization Policy}
\label{app:best_policy}

\begin{figure*}[!t]
    \centering
    \includegraphics[width=\textwidth]{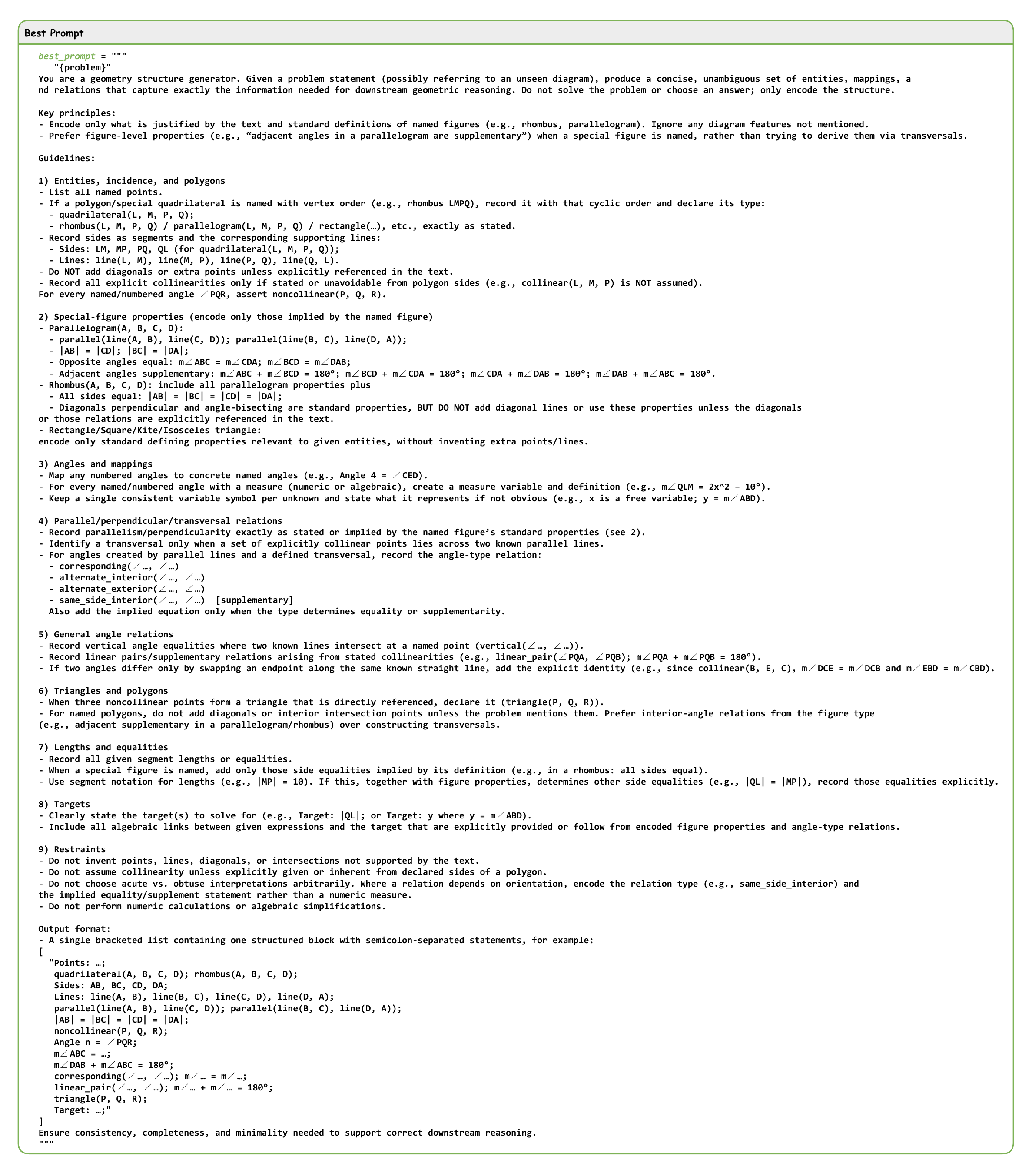}
    \vspace{-4mm}
    \caption{\small Final evolved formalization policy selected by GeoReform.}
    \label{fig:appendix_best_policy_prompt}
    \vspace{-2mm}
\end{figure*}

\end{document}